\documentclass[10pt,twocolumn,letterpaper]{article}

\usepackage{cvpr}              

\usepackage{graphicx}
\usepackage{amsmath}
\usepackage{amssymb}
\usepackage{booktabs}
\usepackage[accsupp]{axessibility}

\usepackage[pagebackref,breaklinks,colorlinks]{hyperref}

\usepackage[capitalize]{cleveref}
\crefname{section}{Sec.}{Secs.}
\Crefname{section}{Section}{Sections}
\Crefname{table}{Table}{Tables}
\crefname{table}{Tab.}{Tabs.}

\def\cvprPaperID{5} 
\def\confName{CVPR}
\def\confYear{2023}

\begin{document}

\title{Training Strategies for Vision Transformers for Object Detection}

\author{Apoorv Singh\\
Motional\\
USA\\
{\tt\small apoorv.singh@motional.com}
}
\maketitle

\begin{abstract}
Vision-based Transformer have shown huge application in the perception module of autonomous driving in terms of predicting accurate 3D bounding boxes, owing to their strong capability in modeling long-range dependencies between the visual features. However Transformers, initially designed for language models, have mostly focused on the performance accuracy, and not so much on the inference-time budget. For a safety critical system like autonomous driving, real-time inference at the on-board compute is an absolute necessity. This keeps our object detection algorithm under a very tight run-time budget. In this paper, we evaluated a variety of strategies to optimize on the inference-time of vision transformers based object detection methods keeping a close-watch on any performance variations. Our chosen metric for these strategies is accuracy-runtime joint optimization. Moreover, for actual inference-time analysis we profile our strategies with float32 and float16 precision with TensorRT module. This is the most common format used by the industry for deployment of their Machine Learning networks on the edge devices. We showed that our strategies are able to improve inference-time by 63\% at the cost of performance drop of mere 3\% for our problem-statement defined in \cref{sec:evaluation}. These strategies brings down Vision Transformers detectors \cite{detr, detr3d, petr, petrv2, bevformer} inference-time even less than traditional single-image based CNN detectors like FCOS\cite{fcos, yolo, ssd}. We recommend practitioners use these techniques to deploy Transformers based hefty multi-view networks on a budge-constrained robotic platform. 
\end{abstract}

\section{Introduction}
\label{sec:intro}
\begin{figure}[t]
  \centering
   \includegraphics[width=\columnwidth]{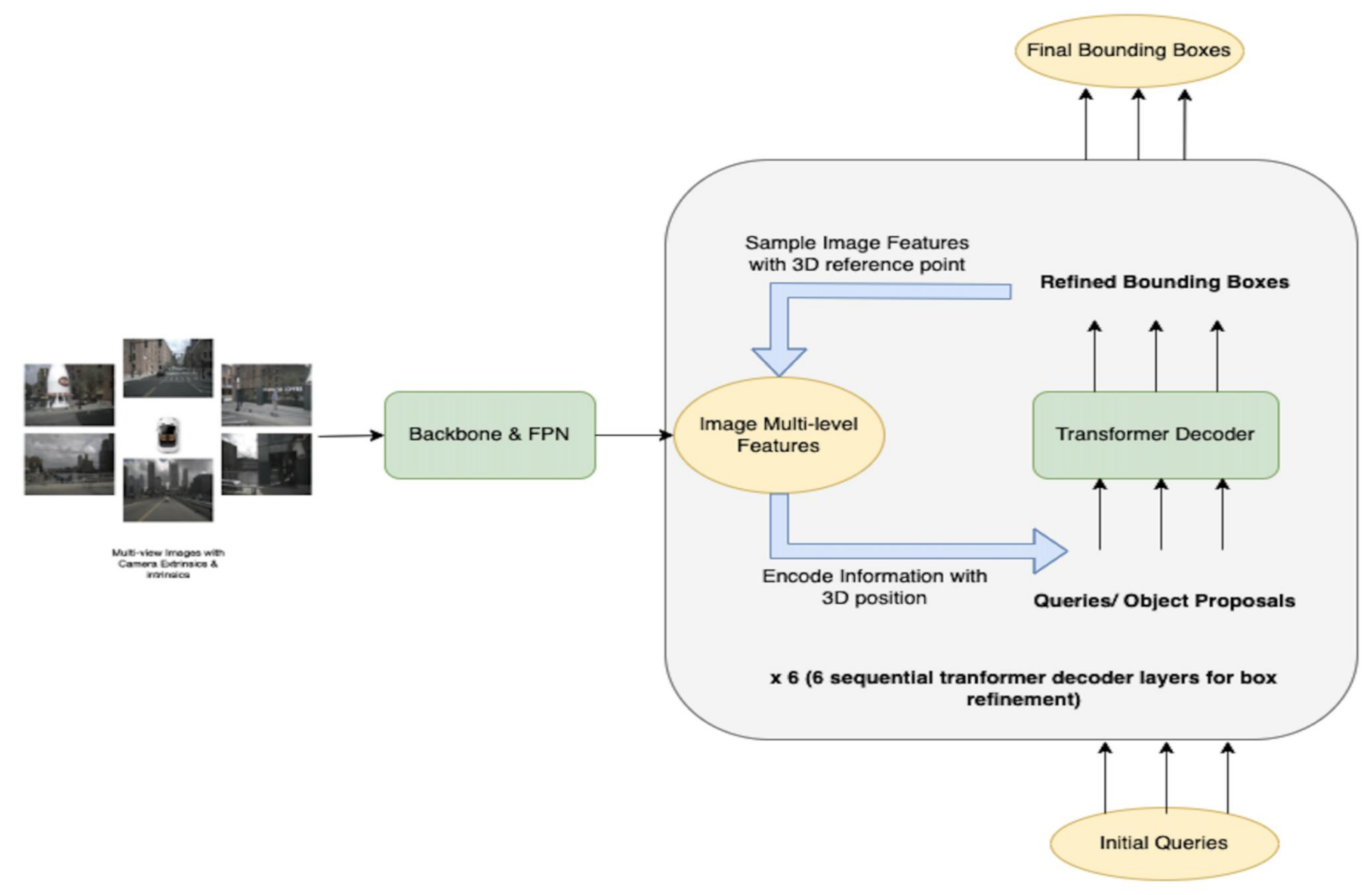}
   \caption{A standard multi-view vision based detector \cite{detr3d} with CNN based backbone and FPN \cite{fpn} and transformers based detection head in gray-box. Inputs: Multi-view images, camera transformation matrices. Outputs: 3D bounding-boxes.}
   \label{fig:detr3d}
\end{figure}
In the last decade, Convolution Neural Networks (CNNs) was driven by the model architectural updates \cite{resnet, se_block, mobilenet, efficientnet} in the field of computer vision. Moreover there have been a plethora of techniques proposed to improve training strategies of these CNN models \cite{review_1, review_2}. Recently, Vision Transformers, first introduced by ViT\cite{vit}, and iteratively reformulated by DETR-based approaches \cite{detr, detr3d, petr, bevformer} has emerged as the better alternative architecture for object detection with images. However, the literature and leader-boards of the Transformers' object-detection community tends to focus most on the architectural impact of these hefty models. When these methods are to be used on an actual robotic-platform, runtime-accuracy joint-optimization is what that matters the most, owing to the fact that any edge-device would have a limited compute-budget. Moreover these ML algorithms have to be operated at high frequency as autonomous cars move fairly fast and need to update their road and dynamic agents' understanding at at-least 10hz (10 times a second). Any top-performing method from a detection leader-board \cite{nuscenes, wod}, would most likely be based out of ensemble approach, which is purely impractical on a compute-budget constrained on-board device. Performance of any machine learning model is dependent on three things: 1) Architecture; 2) Training Strategies; 3) Inference-time budget. This work, unlike others, focuses primarily on the later two: \emph{Training strategies} and \emph{Inference-time budget}. We conclude that with the right network down-scaling strategies we can make Vision Transformers-based detectors practically capable to be fitted onto a deployment platform that operates within the inference-time budget and at high frequency. Transformers, which was initially inspired from language models, had less focus on inference-time bottlenecks early on as most of the processing of language models occurs on cloud the servers for example with ChatGPT. Autonomous car, an extremely safety critical system, has to perform its computation on-board device as we can not live-stream video data from cameras and laser sensors to a cloud servers; have it processed there; and sequentially download the processed predictions back to the device; in addition of doing all that within our real-time operation constraint. 

\subsection{Motivation}
Performance improvements not only come from the novel architectures \cite{detr, detr3d, vit, petr, petrv2, bevformer} but also from the modern scaling strategies of the Vision Transformers network. We noted that this is specially the case for inference-time improvements, as more researchers tend to focus on accuracy compared to inference-time optimizations if at all. Our work is to carefully analyze inference-time optimization strategies that may eventually lead to a Vision-based transformers models practically deploy-able on an autonomous vehicle platforms.

\subsection{Contribution}
Our contributions can be summarized as below: 
\begin{itemize}
    \item We identify modifications in Vision Transformer's architecture of the current State-of-the-art (SoTA) object detection algorithms that lead to better runtime-accuracy optimization. We also list findings on model-precision and training schedule involved in this analysis. As a secondary check of the model we perform MACs (\#operations) analysis to add to the third dimension over inference-time and performance trade-off.
    \item We design post-processing and pre-processing strategies that lead to runtime-performance improvements for Vision Transformers models. 
    \item We design an effective evaluation strategy for an on-car detectors that is able to cater all our accuracy needs and inference-time constraints. We extend inference-time analysis one step further and evaluate the inference-time numbers on a \emph{float16} and \emph{float32} model precision in \emph{TensorRT} model format.
    \item We also discuss further extension ideas for this work that researchers may focus on to further optimize on runtime-accuracy optimization for these models. We link those ideas with the representative work too, bridging a relation from other fields.

Based on our empirical results we claim that carefully adjusting input-image resolution and Transformer-decoder layer embedding space and other parameters can bring down network inference-time down by 63\% with just minimal drop in performance. We evaluate all the results on our in-house dataset which in terms of representation can be compared to any public dataset like nuScenes \cite{nuscenes} or Waymo Open Dataset \cite{wod} but much larger with more diverse classes and scenarios, or in other words much more difficult one too. By adopting proposed strategies, we were able to get inference-time for Vision Transformers detector models close to or in-fact even better than a lot of traditional CNN-based detectors, keeping the detection performance at par with the superior Vision Transformers models. To the author's knowledge this is the first paper which does runtime-accuracy scaling strategies for Vision Transformers models. Moreover we do inference-time analysis on \emph{TensorRT} model format along with MACs analysis.
\end{itemize}
\subsection{Previous Work}
\cite{review_1} paper highlights interesting strategies on model-scaling for CNN models with respect to speed-accuracy Pareto curve. However this analysis lacked runtime analysis on the \emph{TensorRT} model. Moreover this paper focused more on CNN based backbone and not the Transformer based \cite{attention} module which tend to be SoTA detectors in the current literature. \\ \\
\cite{review_2} focuses on the ResNet \cite{resnet} based backbone. They only evaluated their models on a single-image based classification task, ImageNet \cite{imagenet}, which is a fairly simple task compared to multi-view 3D object detection tasks \cite{nuscenes, wod}. Hence a lot of learnings from here might not be valid for on-car deployment of these Transformers based networks. \\ \\
\cite{review_3} focuses on classification performance with \cite{vit} approach, which handled the problem end to end with transformers by dividing patches into 16x16 grids \cite{vit}. This has proven to be a slower and sub-optimal approach for the much more complex problem of multi-view object detection problems. Modern approaches \cite{bevformer, detr3d} have proved to better for this task as per the detection-leaderboard . In addition, this approach focuses on accuracy-training time pareto curve, but our focus is on accuracy-runtime or in other words accuracy-inference time of the network, which is a bigger problem statement in the autonomous vehicle industry. \\ \\
While \cite{review_4} focuses on Transformers but for the application of neural machine translation, which is a significantly different problem compared to the multi-view 3D object detection problem that we are trying to solve with Vision Transformers. \\ \\ 
\cite{scaling_strategies_vit} analysis is probably the most related to our studies however, it is yet different in multiple aspects. Firstly, this scaling studies again focuses on the single-image based classification task of ImageNet \cite{imagenet} and does not have any mobile robot application. This paper also focuses on ViT \cite{vit} based approach, which is an end-to-end transformers based network and not DETR based approaches \cite{detr, detr3d, petr, petrv2, bevformer} that have recently proven better track record in doing 3D object detection for complex datasets of autonomous driving with multi-view cameras. \\ \\
In contrast to the other work, we focus on down-scaling of model with the close-watch on model's performance degradation focused on the Vision Transformers models that are based on CNNs for extracting image features and transformers for detection head to predict boxes in the Bird's Eye View space. Other transformers work {\cite{detr, deformabledetr}} primarily focuses on the architectural impact of their approach, however, we focus on training strategies that lead to joint optimization of performance and run-time thereby making these networks deploy-able on an autonomous vehicle under run-time and compute constraint.

\section{Object Detection Methods}
\label{sec:object_detection}
In this section we will cover different strategies used in object detection starting off with \emph{Single-image Based Detection} and then focusing majorly on \emph{Multi-image Based Detection}.
\subsection{Single-image Based Detection}
Single-image based object detection can be divided into \emph{two-stage}, \emph{single-stage} and \emph{set-based} detectors in terms of chronological invention of these detectors. Two-stage detectors \cite{fasterrcnn, maskrcnn, singh_5} are a class of detectors that are divided into two stages. First stage is to predict arbitrary number of object proposals, and then in second stage they predict boxes by classifying and localizing those object proposals. However, these models suffers through high inference-time bottleneck because of the two-stage nature with redundant computations. Then came the single-stage detectors \cite{yolo, ssd, fcos, centernet} which brings down the inference-time of two-stage detectors. These models either uses heuristics based anchor boxes \cite{ssd, yolo} or center-based heatmap on features \cite{fcos, centernet, fcos3d} to make predictions. However these approaches still relies on limited receptive field associated with the CNNs and fail to develop long-range relations between the image features. This becomes especially the concern for multi-image detection problem in autonomous driving where 6-8 cameras are used to capture the entire $360^{\circ}$ surrounding scene. Next, researchers became interested in bringing transformer-based architecture leanings from language models to the computer vision task. \cite{vit} reformulated the entire CNN-based problem to transformers based problem. This architecture has transformers based encoder a.k.a backbone as well as transformers based detection head. It suffered through inference-time constraints as attention module was applied to even very low-level pixels information and hence shooting up the computation requirement. However, some strategies \cite{scaling_strategies_vit} have been explored on accuracy-runtime pareto curve, but it only focused on relatively easier image classification problem. Subsequently most recent detection work has been focused on CNN-based backbone and transformers based detection head in DETR-based approaches \cite{detr, petr, petrv2, deformabledetr}. These approaches were claimed to be much superior in making sparse predictions in the scene by leveraging self-reasoning of the object-proposals in self-attention layer \cite{attention}. These object-proposals are later refined as a set of predictions of boxes. This approach has also extended its way to the multi-view object detection problem as discussed in the next section.
\subsection{Multi-image Based Detection}
For a lot of robotics applications including autonomous driving, perception module needs to make prediction for the entire $360^{\circ}$ scene \cite{singh_1}, and that too in the Bird's eye view (BEV) so that it can be easily consumed by the other autonomy's software module like path-prediction and path-planning. Until the last few years, multi-view based detection problem was handled using per-view Machine learning based methods and then all the detections were consolidated using heuristic methods of data association using IoU (Intersection over Union). This post-processing step merges duplicate detections coming from the adjacent camera-views. Recently a lot of work have been explored in learning end-to-end differentiable network for multi-view detection. These approaches can be broadly classified as per \cite{singh_1} by (1) \emph{Geometry based view transformers} and (2) \emph{Cross-attention based vision-transformer}. Geometry based models are typically based out of \cite{lss} approach, which as an intermediate step creates BEV pseudo point-cloud by discretizing depth for each pixel and then runs a LiDAR based detector \cite{centernet} on the pseudo point-cloud frustum from all the cameras. Due to the inherent two-stage process these models tend to have huge inference-time, despite being entirely based out of CNNs. Cross-attention based vision-transformer based approaches \cite{detr3d, bevformer} have shown to be SoTA in the leader-boards \cite{nuscenes, wod}. These approaches typically have CNN based backbone and transformers based head which feeds in queries and CNN features to predict 3D bounding boxes \cite{singh_2, singh_3, singh_4}. \cite{detr3d, petr, petrv2} have focused their work on sparse-queries that are either learned from the training-dataset or constructed using input-data, whereas \cite{bevformer} focuses on dense queries in the BEV map. For the scope of this paper we focus our methods on sparse-query based approaches which are much more optimal in terms of inference-time. Swin-transformer \cite{swin} has explored in converting CNN based backbone to transformers based one using hierarchical Transformer whose representation is computed with shifted windows.

\begin{figure}[t]
  \centering
   \includegraphics[width=\columnwidth]{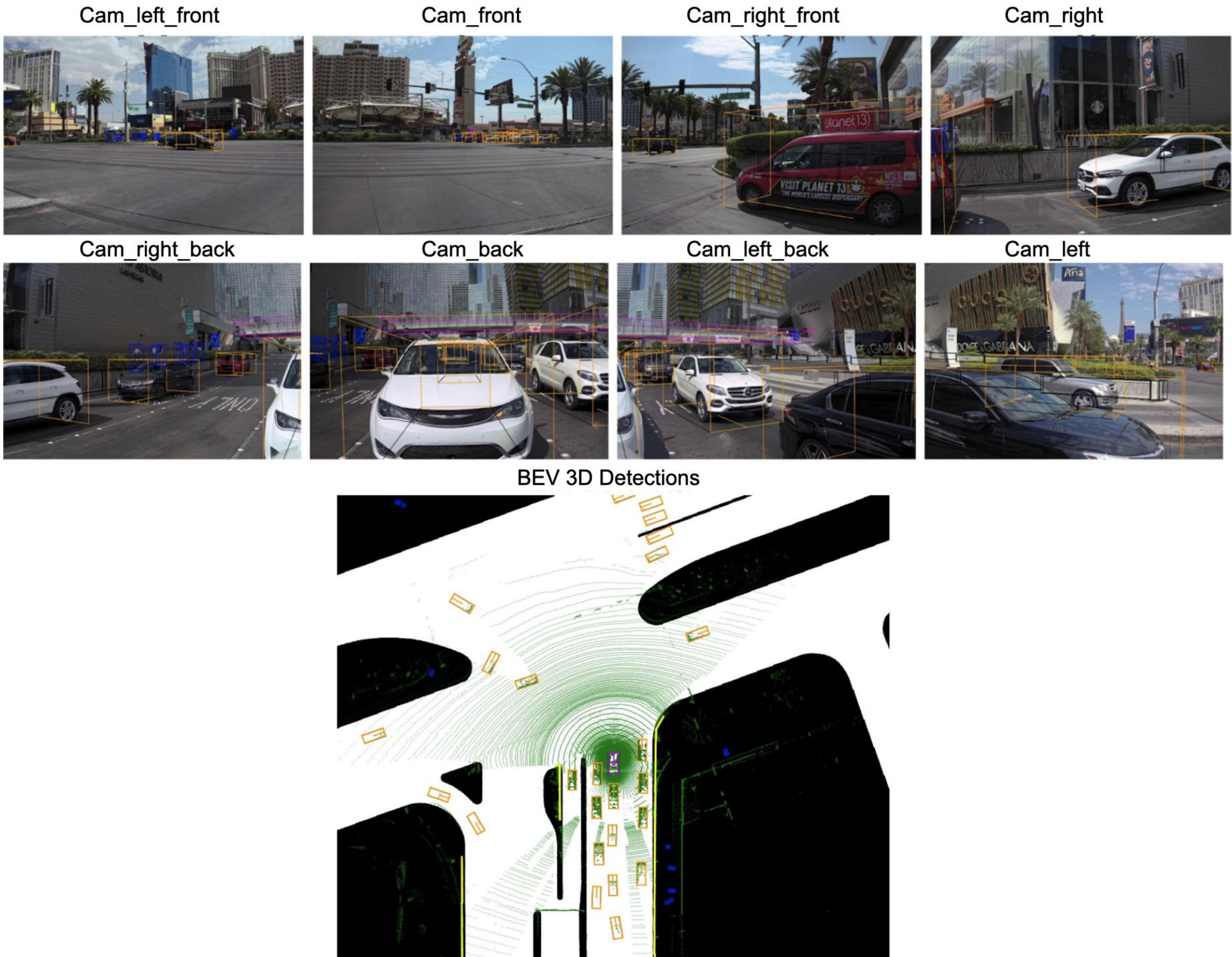}
   \caption{Input-Output of the Problem Statement. Surround-view 8 camera images (top); LiDAR Point Cloud overlayed over an HD Map (bottom). Key: Green points: LiDAR point cloud; Pink box: Autonomous vehicle; Black-white map: Pre-computed HD map.}
   \label{fig:bev_output}
\end{figure}

\section{Evaluation Criteria}
\label{sec:evaluation}
In this section we discuss different fronts used for comparative analysis for different models viz., \emph{Accuracy}, \emph{Inference-time a.k.a run-time} and few less used ones: \emph{Multiply–accumulate operation (MACs)} and {Number of Parameters}.
\subsection{Model Accuracy}
3D object detectors use multiple criteria to measure performance of the detectors viz., precision and recall. However, mean Average Precision (mAP) is the most common evaluation metric. Intersection over Union (IoU) is the ratio of the area of overlap and area of the union between the predicted box and ground-truth box. An IoU threshold value (generally 0.5) is used to judge if a prediction box matches with any particular ground-truth box. If IoU is greater than the threshold, then that prediction is treated as a True Positive (TP) else it is a False Positive (FP). A ground-truth object which fails to detect with any prediction box, is treated as a False Negative (FN). Precision is the fraction of relevant instances among the retrieved instances; while recall is the fraction of relevant instances that were retrieved.
\begin{equation}
Precision=TP/(TP+FP)
\end{equation}
\begin{equation}
Recall=TP/(TP+FN)
\end{equation}
Based on the above equations, average precision is computed separately for each class. To compare performance between different detectors (mAP) is used. It is a weighted mean based on the number of ground-truths per class. Alternatively F1 score is the second most common detection metric, which is defined as weighted average of the precision and recall. Higher \emph{AP} detectors gives better performance when the model is deployed at varied confidence threshold, however higher \emph{max-F1} score detector is used when the model is to be deployed at a known fixed optimal-confidence threshold score. 
\begin{equation}
F1=2*Precision*Recall/(Precision+Recall)
\end{equation}
For all our evaluations we restrict our Ground-truths and predictions up-to \emph{60meters} in the Bird's eye view space from the ego-vehicle for comparison at common-grounds.

\subsection{Inference-time}
In-terms of practical application, another equally important metric for detectors is the inference-time, defined as run-time during deployment of the model. We compare our models on \emph{PyTorch} model at standard precision \emph{float32} and also at optimized precision of \emph{float16} in \cref{sec:model_precision}. In most of the robotic applications deployment platforms is \emph{Nvidia GPU}, for which \emph{TensorRT} is used as the deployment library. \emph{TensorRT} runtimes and \emph{PyTorch} module inference-time aren not always correlated. So it is not a fair comparison of inference-time of the model until we do this analysis with the actual \emph{TensorRT} format. In addition to \emph{TensorRT float32} model precision we show run-time comparison in \emph{TensorRT float16} format a.k.a \emph{half-precision} as well. 
\subsection{MACs and Parameters}
MACs (Multiply–accumulate operation) and number of parameters in any ML model are other less used metrics to compare two models size along with the inference-time. While inference-time of the network is dependent on the hardware but MACs and number of parameters are hardware-agnostic parameters. MACs defined number of computations happen during inference of the model. These computations typically include number of \emph{multiply} and \emph{add} operations on the float numbers. These metrics are generally only used when deployment hardware is not known while development of the ML network. MACs and parameters might not always correlate with inference-time, which is more predictive measure of latency of the network. 

\section{Methodology and Experiments}
In this section we will present series of network down-scaling methods. We train our model on our in-house autonomous driving dataset with 3D BEV boxes as annotations and evaluated as per \emph{mAP} and \emph{max-F1} score mentioned in \cref{sec:evaluation} and inference time calculated on a \emph{T4 Nvidia GPU} in different model formats. All the models are trained and evaluated on the same GPU as well. One sample of our dataset is shown in \cref{fig:bev_output}. Our training strategies can be categorized as: \emph{Pre-training Strategies}, \emph{Training Strategies}, \emph{Post-training Strategies}, \emph{Model Formats and Precision}. For profiling measures: inference-time is measure in milliseconds, MACs are measured as number of operations and Paramters are measured as number of parameters.
\subsection{Pre-training Strategies}
\label{sec:pre-training}
\subsubsection{Input Resolution}
To understand the over-head of backbone and detection head as shown in \cref{fig:detr3d}, we ran profiling with inference-time, parameters count and mac operations on it. Results from \cref{tab:profiling} shows that most of the inference-time overhead lies in the backbone. We deduced that image being a 2D matrix; total inference-time can be quadratically reduced if we can reduce the input resolution keeping a close watch on accuracy performance change. We trained and evaluated our model by aggressively reducing the input resolution as shown in \cref{tab:image-resolution}. We noticed that by reducing image-resolution smaller objects like pedestrians is affected more adversely compared to larger object vehicles. We concluded that we can safely reduce down the resolution to $(598, 394)$ i.e. $3/5^{th}$ of the original ($996, 656)$ with $52.5\%$ inference-time improvement at the cost of mere $1.4\%$ drop in the pedestrian AP (Average precision) performance and the $0.5\%$ drop in the vehicle AP performance. Also note that the number of parameters don't change with input-resolution as it is based out of fully-convolutional based backbone\cite{se_block} and an FPN \cite{fpn}. For further experiments we will focus more on the inference-time metric rather than MACs and Parameters, as inference-time is a more representative metric for the given hardware. For further experiments we will change our baseline to $(598, 394)$ input-resolution to show accumulative effect of our strategies.

\begin{table}
  \centering
  \begin{tabular}{@{}lccr@{}}
    \toprule
    Module & Inference-time & MACs  & Parameters \\
    \midrule
    Entire Network  & 447 & 832B & 37M \\
    Backbone + FPN  & 380 & 829B & 34M \\
    Detection Head  & 67 & 2B & 3M \\
    \bottomrule
  \end{tabular}
  \caption{Profiling different stages of the network in terms of inference-time, MACs and number of parameters.}
  \label{tab:profiling}
\end{table}

\begin{table*}
  \centering
  \begin{tabular}{@{}lccccr@{}}
    \toprule
    Input-resolution & Vehicle (AP) & Pedestrian (AP) & Inference-time (ms) & MACs  & Parameters \\
    \midrule
    996, 656  & 64.9 & 71.8 & 447 & 832B & 37M \\
    664, 437  & 64.5 & 71.5 & 242 & 372B & 37M \\
    \textbf{598, 394}  & 64.0 & 71.4 & 212 & 305B & 37M \\
    498, 328  & 60.4 & 70.6 & 166 & 212B & 37M \\
    \bottomrule
  \end{tabular}
  \caption{Effects of reducing input-resolution on accuracy and inference-time. Input-resolution is represented as pixels in width and height dimension respectively.}
  \label{tab:image-resolution}
\end{table*}

\subsubsection{Image Pre-croppers}
In object detection, image Pre-cropping refers to cropping of the image from certain edges of the camera which represents least informative pixels to squeeze out model inference-time from it by saving onto those pixels' computations. To select optimal pre-cropping strategy for our current network we used two methods, firstly we visually analyzed images as shown in \cref{fig:bev_output} and tried to find the least informative part of the image in-terms of our problem statement i.e. detect all the safety critical agents on the road. We noticed that generally top few-tens of pixels in the image are the least important as they mostly include top of a tree/ top of a building/ top of a pillar etc. as shown in \cref{fig:bev_output}. Our second methods was quantitative analysis in which we gathered all the boxes that lies in those cropped top few-tens pixels to confirm that we are not pruning information of any of our objects of interest with those cropped pixels. We found that less than $0.25\%$ of our ground-truth in training dataset is actually in the top-50 pixels for input image-resolution of $(598, 394)$. Our empirical findings are shown in \cref{tab:precrop}. We noticed that removing pixels from the top has no/ positive effect on pedestrians and only minor effect on vehicles, as some of the features of vehicles might get missed specially for a very close-range vehicles which occupy the entire image. We concluded that removing 50 pixels from top is an optimal strategy for inference-time and accuracy joint-optimization with 10\% improvement in inference-time at the cost of only 2\% accuracy drop in vehicles and no drop in pedestrian.
\begin{table*}
  \centering
  \begin{tabular}{@{}lcccr@{}}
    \toprule
    Input-resolution & Cropped pixels (Top) & Pedestrian (AP)  & Vehicle (AP) & Inference-time \\
    \midrule
    598, 394 & 0  & 64.0 & 71.4 & 212 \\
    598, 394 & 25  & 64.8 & 70.3 & 200 \\
    598, 394 & 50  & 64.8 & 69.6 & 191 \\
    598, 394 & 100  & 62.8 & 65.4 & 175 \\
    \bottomrule
  \end{tabular}
  \caption{Empirical experiment with Pre-croppers.}
  \label{tab:precrop}
\end{table*}

\subsection{Training Strategies}
\label{sec:training}
\subsubsection{Training Schedule}
For our baseline experiments we used one-cycle learning rate (L.R.) scheduler \cite{one-cycle} with maximum learning rate as $5e^{-5}$. We experimented with three values associated with one-cycle learning rate viz., Start L.R; Max L.R. and End L.R. Empirical results are shown in \cref{tab:lr}. We concluded that there's a value in using one-cycle learning rate with its limits by using higher learning rate in the middle of training and relatively much smaller learning rate at the beginning and end of the training. We also observed if we try to increase the ending learning rate, we suffer through training-divergence. 
\begin{table}
  \centering
  \begin{tabular}{@{}lccccr@{}}
    \toprule
    \small{L.R. Star} & \small{L.R. Max.} & \small{L.R. End} & \small{Pedestrian} & \small{Vehicle} \\
    \midrule
    $5e^{-6}$ & $5e^{-5}$ & $5e^{-6}$ & 60.7 & 67.0 \\
    $1e^{-5}$ & $1e^{-4}$ & $1e^{-5}$ & 63.1 & 72.4 \\
    $1e^{-6}$ & $1e^{-4}$ & $1e^{-5}$ & 63.4 & 70.8 \\
    $5e^{-6}$ & $5e^{-4}$ & $5e^{-6}$ & 64.0 & 72.9 \\
    \bottomrule
  \end{tabular}
  \caption{Empirical experiment with the one-cycle learning rate scheduler. Pedestrian and vehicle performance is measured in AP (Average Precision).}
  \label{tab:lr}
\end{table}
\subsubsection{Number of Decoders}
Transformers decoders are sequential transformer layers that iteratively refines object-proposals so that they eventually converge to predictions that match the ground-truth as defined in \cite{detr}. In this section, we try to build a relation on how many number of sequential decoder layers are optimal. Similar analysis has also been done in \cite{detr3d}, but we wanted to confirm the findings based on our own dataset and other training conditions. With our empirical experiments as shown in \cref{tab:decoder_num}, we concluded that our baseline number of decoder layers i.e. \textbf{6}, is indeed the most optimal number of decoders for the problem statement. 
\begin{table}
  \centering
  \begin{tabular}{@{}lccr@{}}
    \toprule
    \#Decoders & Pedestrian  & Vehicle & Inference-time \\
    \midrule
    1  & 40.6 & 49.7 & 134 \\
    2  & 57.9 & 64.3 & 144 \\
    3  & 60.1 & 66.1 & 155 \\
    5  & 61.9 & 67.4 & 179 \\
    6  & 64.8 & 69.6 & 191 \\
    7  & 65.0 & 70.1 & 202\\
    \bottomrule
  \end{tabular}
  \caption{Empirical experiment with number of transformers decoders. Pedestrian and vehicle performance is measured in AP (Average Precision).}
  \label{tab:decoder_num}
\end{table}

\subsubsection{Embedding Dimensions}
In this section we try to formulate the relevance of the long embedding dimension of the queries. Queries are latent representation of the object-proposals which are refined using transformers-decoders for making final predictions. Visualizations of the query in image-space can be referred in \cite{detr}. Baseline taken from \cite{detr, detr3d} has embedding dimension of $256$, but based on our results shown in \cref{tab:embedding_dim}, $128$ dimension gave $9.5\%$ better inference-time with no performance drop at all. It is also worth noting that due to internal optimization of the GPU, embedding dimension has to be of the power of 2 for inference-time optimization. This can be empirically seen in the row with $96$ embedding dimension where inference-time is close to the one with $128$ dimension-size despite having less computations. 
\begin{table}
  \centering
  \begin{tabular}{@{}lccr@{}}
    \toprule
    \small{Embedding Dimension} & \small{Pedestrian}  & \small{Vehicle} & \small{Inference-time} \\
    \midrule
    256  & 64.8 & 69.6 & 211 \\
    128  & 64.7 & 69.9 & 191 \\
    96  & 64.2 & 68.4 & 190 \\
    64  & 61.4 & 66.7 & 183 \\
    \bottomrule
  \end{tabular}
  \caption{Empirical experiment with the varied embedding dimensions of the decoder. Note: FPN \cite{fpn} and transformer queries dimensions are changed together in these experiments. Pedestrian and vehicle performance is measured in AP (Average Precision).}
  \label{tab:embedding_dim}
\end{table}
\subsubsection{Number of Queries}
In this section we experimented with the number of object-proposals a.k.a queries in transformers head. This parameter is specially very sensitive to Vision transformer head because we natively don't perform Non-maximum suppression on the detection output, owing to the fact of using set-based loss. Set-based loss theoretically forces to make only single prediction per ground-truth object, in-comparison to dense-prediction methods \cite{yolo, fcos, ssd, bevformer} which makes multiple prediction per ground-truth. Theoretical way to get the optimal number for queries is by analyzing the training dataset and finding average number of ground-truths present per training sample. Then we may add $10-20\%$ of that number as a buffer for negative class mining. In addition to these analysis we empirically show results in \cref{tab:query_num}. On an average our training dataset has ~300 agents per $360^{\circ}$ view. We concluded that using $400$ queries are optimal for this problem-statement, with a $5\%$ inference-time improvement at the cost of no performance. Moreover we actually noticed that precision has increased with $400$ query model when compared to $900$ queries because of more relevant predictions. 
\begin{table}
  \centering
  \begin{tabular}{@{}lccr@{}}
    \toprule
    \#Queries & Pedestrian  & Vehicle & Inference-time \\
    \midrule
    900  & 64.8 & 69.6 & 191 \\
    700  & 64.6 & 69.5 & 187 \\
    500  & 65.0 & 69.9 & 183 \\
    400  & 64.8 & 70.0 & 181 \\
    300  & 61.5 & 66.7 & 181 \\
    \bottomrule
  \end{tabular}
  \caption{Empirical experiment with the number of queries. Pedestrian and vehicle performance is measured in AP (Average Precision).}
  \label{tab:query_num}
\end{table}
\subsection{Post-training Strategies}
In this study we go through different post-processing techniques we can use to maximize our performance once the training has completed. These post-processing strategies' inference-time overhead can be treated as negligent compared to the model inference-time. 
\label{sec:post-training}
\subsubsection{Pick Top-k boxes}
In baseline paper \cite{detr3d} we noted that the authors have made use of top-k boxes to filter predictions based on confidence values. We experimented by removing this top-k logic with a parameter sweep, but we concluded that having top-k boxes as 90\% of the boxes a.k.a number of queries seemed to be the optimal number with vision transformer detectors.

\subsubsection{Non-Maximum Suppression (NMS)}
Seminal paper \cite{detr, detr3d, petr}, claimed against using NMS approach with transformers based network. Their rationale behind this claim was that self-attention layer and set-based loss during training of the network are enough to not let the network make duplicate detections for an object. However, even after tuning down number of queries to just $400$, we noticed that there are some overlap bounding boxes for a single object. This is a clear application of the NMS method. We went ahead and implemented a simple version of class-agnostic NMS to see performance difference with the baseline in \cref{tab:nms}. We concluded that NMS has clear gain in vehicle class, as these boxes are generally bigger and transformers may try to associate a ground-truth with multiple queries. However, performance improvement wasn't very noticeable in case of smaller sized class i.e. pedestrians.
\begin{table}
  \centering
  \begin{tabular}{@{}lccr@{}}
    \toprule
    NMS & IoU-threshold  & Pedestrian & Vehicle \\
    \midrule
    $\times$  & - & 64.0 & 72.9 \\
    $\surd$  & 0.5 & 63.1 & 72.5 \\
    $\surd$  & 0.3 & 63.5 & 72.6 \\
    $\surd$  & 0.2 & 64.2 & 73.6 \\
    \bottomrule
  \end{tabular}
  \caption{Empirical experiment with NMS post-processing. Pedestrian and vehicle performance is measured in AP (Average Precision).}
  \label{tab:nms}
\end{table}

\subsection{Model Formats and Precision}
\label{sec:model_precision}
In this section we will highlight inference-time improvements we can get with the hardware accelerators after reducing the precision of the models. We will cover two formats: \emph{PyTorch} and \emph{TensorRT} with two different model precision: \emph{float32} and \emph{float16} in \cref{tab:precision}. For \emph{PyTorch} model format we noted there's approximate $50\%$ inference-time boost just using half-precision a.k.a \emph{float16} model precision. In addition to that \emph{TensorRT} has further speedups by merging layers together and hence reducing number of operations. \emph{int8} model precision wasn't studied in this analysis. It may provide further boost in the inference-time of these models. 
\label{sec:model-format}

\begin{table}
  \centering
  \begin{tabular}{@{}lcr@{}}
    \toprule
     &  PyTorch & TensorRT  \\
    \midrule
    float32  & 187 & 148 \\
    float16  & 98 & 62 \\
    \bottomrule
  \end{tabular}
  \caption{Model precision and model-format inference-time comparison. All numbers are measured on the same T4 Nvidia GPU.}
  \label{tab:precision}
\end{table}

\section{Discussions}
\label{sec:Discussions}
\subsection{Societal Impact}
We ran a series of scaling experiments to derive these training strategies on multiple-multi GPU machines for multiple days. Eventually we were able to prove our hypothesis on down-scaling strategies for hefty vision transformers models. We are hopeful that the cost and environmental effect associated with these experiments would be much lesser compared to the long term benefits these learning would have for future researchers and deployment engineers. 
\subsection{Further Extensions}
\label{sec:extension}
Here we will cover the aspects which were not in the scope of the paper but authors believe would be worthwhile to explore for future Vision Transformers strategies. Whole lot of work has happened over CNN-based Neural Architecture Search (NAS) \cite{rigging, lth} but we haven’t seen any of such work associated with Vision Transformer models yet. It could be a great direction to explore and effectively convenient too as transformers tend to have less hyper-parameters compared to the CNN networks to search for. In addition we can develop model pruning strategies for Vision Transformers models, looking at inference-time improvements with CNN based Vision models \cite{prune_1, prune_2}, we might see some low hanging fruits here. Lastly we think for papers with dense queries \cite{bevformer}, researchers could develop on smart ways to automatically block some of the non-practical queries based on the prior HD-map or physics modeling etc. to get gain back the low-latency model.

\section{Conclusion}
\label{sec:conclusion}
In this work, we analyze runtime-accuracy performance numbers on different network related and pre-processing and post-processing related strategies on Vision Transformers model. We took an extra step of also comparing model runtimes on \emph{TensorRT} modules in \emph{float32} and \emph{float16} precision which is more of a standard practice in the autonomous industry. In addition we also did MACs and \#parameters analysis with these techniques to understand differences in the model as per the operations number. We concluded that depending on the problem statement, bigger and deeper models do not always lead to better results; despite wasting resources and environmental impact. With all our strategies, we are able to improve inference-time by $63\%$ at the cost of performance drop of mere $3\%$ for our problem-statement. We hope researchers and engineers community would benefit from these deployment-time analysis of Vision transformers models on a robotic platforms and can unblock some of the concerns we are facing for the path to deployment for these models. Also we hope to get some environmental positive impact towards scaling down energy-hungry compute resources for beefier object detection algorithms.

{\small
\bibliographystyle{ieee_fullname}
\bibliography{egbib}
}

\end{document}